\documentclass[fleqn,10pt]{wlscirep}
\usepackage[utf8]{inputenc}
\usepackage[T1]{fontenc}
\usepackage{cite}
\usepackage{amsmath}
\usepackage{amssymb}
\usepackage{algorithmic}
\usepackage{url}
\usepackage{graphicx}
\usepackage{epsfig}
\usepackage{soul}
\usepackage{color}
\usepackage{amsmath}
\usepackage{amssymb}
\usepackage{xcolor}
\usepackage{booktabs}
\usepackage{multirow}
\usepackage[T1]{fontenc}
\usepackage[utf8]{inputenc}
\usepackage{babel}
\usepackage[font=small,labelfont=bf]{caption}

\title{Deep Radiomics for Brain Tumor Detection and Classification from Multi-Sequence MRI}
\author[1,2, *]{Subhashis~Banerjee}
\author[1]{Sushmita~Mitra}
\author[3]{Francesco~Masulli}
\author[3]{Stefano~Rovetta}
\affil[1]{Indian Statistical Institute, Machine Intelligence Unit, Kolkata, 700108, India}
\affil[2]{University of Calcutta, Department of Computer Science and Engineering, Kolkata, 700106, India}
\affil[3]{University of Genova, Dept of Informatics Bioengineering Robotics and Systems Engineering, Genoa, 16146, Italy}

\affil[*]{mail.sb88@gmail.com}

\begin{abstract}
Glioma constitutes $80\%$ of malignant primary brain tumors in adults, and is usually classified as High Grade Glioma (HGG) and Low Grade Glioma (LGG). The LGG tumors are less aggressive, with slower growth rate as compared to HGG, and are responsive to therapy. Tumor biopsy being challenging for brain tumor patients, noninvasive imaging techniques like Magnetic Resonance Imaging (MRI) have been extensively employed in diagnosing brain tumors. Therefore, development of automated systems for the detection and prediction of the grade of tumors based on MRI data becomes necessary for assisting doctors in the framework of augmented intelligence. In this paper, we thoroughly investigate the power of Deep Convolutional Neural Networks (ConvNets) for classification of brain tumors using multi-sequence MR images. We propose novel ConvNet models, which are trained from scratch, on MRI patches, slices, and multi-planar volumetric slices. The suitability of transfer learning for the task is next studied by applying two existing ConvNets models (VGGNet and ResNet) trained on ImageNet dataset, through fine-tuning of the last few layers. Leave-one-patient-out (LOPO) testing, and testing on the holdout dataset are used to evaluate the performance of the ConvNets. Results demonstrate that the proposed ConvNets achieve better accuracy in all cases where the model is trained on the multi-planar volumetric dataset. Unlike conventional models, it obtains a testing accuracy of $95\%$ for the low/high grade glioma classification problem. A score of $97\%$ is generated for  classification of LGG with/without 1p/19q codeletion, without any additional effort towards extraction and selection of features. We study the properties of self-learned kernels/ filters in different layers, through visualization of the intermediate layer outputs. We also compare the results with that of state-of-the-art methods, demonstrating a maximum improvement of $7\%$ on the grading performance of ConvNets and $9\%$ on the prediction of 1p/19q codeletion status.    
\end{abstract}
\begin{document}

\flushbottom
\maketitle

\section*{Introduction}
{M}agnetic Resonance Imaging (MRI) has become the standard non-invasive technique for brain tumor diagnosis over the last few decades, due to its improved soft tissue contrast \cite{deangelis2001brain, cha2006update}. Gliomas constitute $80\%$ of all malignant brain tumors originating from the glial cells in the central nervous system. Based on the aggressiveness and infiltrative nature of the gliomas the World Health Organization (WHO) broadly classified them into two categories, viz. Low-grade gliomas (LGG), consisting of low-grade and intermediate-grade gliomas (WHO grades II and III), and high-grade gliomas (HGG) or glioblastoma (WHO grade IV) \cite{louis16}. Diffuse LGG are infiltrative brain neoplasms which include histological classes astrocytomas, oligodendrogliomas, and oligoastrocytomas and World Health Organization (WHO) grade II and III neoplasms \cite{louis16}. Although LGG patients have better survival than those with HGG, the LGGs are found to typically progress to secondary GBMs and eventual death \cite{liyun13}. In both cases a correct treatment planning (including surgery, radiotherapy, and chemotherapy separately or in combination) becomes necessary, considering that an early and proper detection of the tumor grade can result in good prognosis \cite{van2012adjuvant}.

Histological grading, based on stereotactic/surgical biopsy test, is primarily used for the management of gliomas. Typically the highest grade component, among the histopathology samples obtained, is used to predict the overall tumor grade. Gliomas being heterogeneous, sometimes histopathology samples collected from different parts of the same tumor exhibit different grades. Since pathologists are not provided with the entire delineated tumor during examination, it is likely that the highest grade component may be missing in the biopsy sample. This is called the biopsy sampling error \cite{Chandrasoma89Stereotactic,glantz1991influence,jackson2001limitations}, and can potentially result in wrong clinical management of the disease. Moreover there exist several risk factors in the biopsy test, including bleeding from the tumor and brain due to the biopsy needle; causing severe migraine, stroke, coma and even death. Other associated risks involve infection or seizures \cite{field2001comprehensive,mcgirt2005independent}.

MR imaging, on the other hand, has the advantage of being able to scan the entire tumor in vivo and can demonstrate a strong correlation with histological grade. It is also not susceptible to sampling error, and inter- and intra-observer variability. In this context multi-sequence MRI plays a major role in the detection, diagnosis, and management of brain cancers in a non-invasive manner. Recent literature reports that computerized detection and diagnosis of the disease, based on medical image analysis, could be a good alternative. Decoding of tumor phenotype using noninvasive imaging is a recent field of research, known as {\it Radiomics} \cite{mitra2015medical, mitra2014integrating, gillies2015radiomics}, and involves the extraction of a large number of quantitative imaging features that may not be apparent to the human eye. An integral part of the procedure involves manual or automated delineation of the 2D region of interest (ROI) or 3D volume of interest (VOI) \cite{banerjee2016single, banerjee2016novel, banerjee2018automated, somosmitaPlos2017}, to focus attention on the malignant growth. This is typically followed by the extraction of suitable sets of hand-crafted quantitative imaging features from the ROI or VOI, to be subsequently analyzed through machine learning towards decision-making. Feature selection enables the elimination of redundant and/or less important subset(s) of features, for improvement in speed and accuracy of performance. This is particularly relevant for high-dimensional radiomic features, extracted from medical images. 

Quantitative imaging features, extracted from MR images, have been investigated in literature for the assessment of brain tumors \cite{zhou2017radiomics, gillies2015radiomics}. Ref. \cite{banerjee2017fuzzieee} presents an adaptive neuro-fuzzy classifier, based on linguistic hedges (ANFC-LH), for predicting the brain tumor grade using $56$ 3D quantitative MRI features extracted from the corresponding segmented tumor volume(s). Quantitative imaging features, extracted from pre-operative gadolinium-enhanced T1-weighted MRI, were investigated for the diagnosis of \textit{meningioma} grades \cite{coroller2016d}. A study of MR imaging features was made \cite{zhao2014can} to determine those which can differentiate among grades of \textit{soft-tissue sarcoma}. The features investigated include signal intensity, heterogeneity, margin, descriptive statistics, and perilesional characteristics on images, obtained from each MR sequence. Brain tumor classification and grading, based on 2D quantitative imaging features like texture and shape (involving gray-level co-occurrence, run-length, and morphology), were also reported \cite{zacharaki2009classification}.

Although the techniques demonstrate good disease classification, their dependence on hand-crafted features requires extensive domain knowledge, involves human bias, and is problem-specific. Manual designing of features typically requires greater insight into the exact characteristics of normal and abnormal tissues, and may fail to accurately capture some important representative features; thereby hampering classifier performance. The generalization capability of such classifiers may also suffer due to the discriminative nature of the methods, with the hand-crafted features being usually designed over fixed training sets. Subsequently manual or semi-automatic localization and segmentation of the ROI or VOI is also needed to extract the quantitative imaging features \cite{banerjee2016novel, banerjee2016single}.

Convolutional Neural Networks (ConvNets) offer state-of-the-art framework for image recognition or classification \cite{he2016deep, lecun2015deep, szegedy2015going}.  ConvNet architecture is designed to loosely mimic the fundamental working of the mammalian visual cortex system. It has been shown that the visual cortex has multiple layers of abstractions which look for specific patterns in the input vision. A ConvNet is built upon a similar idea of stacking multiple layers to allow it to learn multiple different abstractions of the input data. These networks automatically learn mid-level and high-level representations or abstractions from the input training data, in the form of convolution filters that are updated during the training process. They work directly on raw input (image) data, and learn the underlying representative features of the input which are hierarchically complex, thereby ruling out the need for specialized hand-crafted image features. Moreover ConvNets require no prior domain knowledge and can automatically learn to perform any task just by working through the training data.

However training a ConvNet from scratch is generally difficult because it essentially requires large training data, along with the significant expertise to select an appropriate model architecture for proper convergence. In medical applications data is typically scarce, and expert annotation is expensive. Training a deep ConvNet requires huge computational and memory resources, thereby making it extremely time-consuming. Repetitive adjustments in architecture and/or learning parameters, while avoiding overfitting, make deep learning from scratch a tedious, time-consuming, and exhaustive procedure. Transfer learning offers a promising alternative, in case of inadequate data, to fine tune a ConvNet pre-trained on a large set of available labeled images from some other category \cite{oquab2014}. This helps in speeding up convergence, while lowering computational complexity during training \cite{tajbakhsh2016, phan2016}.

The adoption rate of ConvNets in medical imaging  has been on the rise \cite{Greenspan2016_guest_survey}. However given the insufficiency of medical image data, it often becomes difficult to use deeper and more complex networks. Application of ConvNets in gliomas have been mostly reported for the segmentation of abnormal regions from 2D or 3D MRIs \cite{Pereira2016, zikic14, Urban2014_brats, kamni17, Davy2015_brats, lyksb15}. Automated detection and extraction of High Grade Gliomas (HGG) was performed using ConvNets \cite{baner18b}. The two-stage approach first identified  the presence of HGG, followed by a bounding box based tumor localization in each ``abnormal'' MR slice. As part of the Computer-Aided Detection system, Classification and Detection ConvNet architectures were employed. Experimental results demonstrated that the  CADe system, when used as a preliminary step before segmentation, can allow improved delineation of tumor region while reducing false positives arising in normal areas of the brain. Recently Yang \textit{et al.}\cite{yang2018glioma} explored the role of deep learning and transfer learning for accurate grading of gliomas, using conventional and functional MRIs. They used a private Chinese hospital database containing $113$ pathologically confirmed glioma patients, of which there were $52$ LGG and $61$ HGG samples. The AlexNet and GoogLeNet were trained from scratch, and fine-tuned from models that had been pre-trained on the large natural image database ImageNet. Testing on the $20\%$ heldout data, randomly selected at patient-level, resulted in maximum test accuracy ($90\%$) by GoogLeNet. Radiomics has also been employed \cite{cho2018classification} for grading of gliomas into LGG and HGG, with the MICCAI BraTs 2017 dataset\cite{bakas2017advancing} being used for training and testing of models.

In addition to tumor grading, the prediction of 1p/19q codeletion status serves as a crucial molecular biomarker towards prognosis in LGG. It is found to be related to longer survival, particularly for oligodendrogliomas which are more sensitive to chemotherapy. Such noninvasive prediction through MRI can, therefore, lead to avoiding invasive biopsy or surgical procedures. Predicting 1p/19q status in LGG from MR images using ConvNet was reported \cite{akkus2017predicting}. The network was first trained on a brain tumor patient database from the Mayo Clinic, containing a total of $159$ LGG cases ($57$ non-deleted and $102$ codeleted) and having preoperative postcontrast-$T1$ and $T2$ images\cite{erickson2017data}. The model was also trained and tested on $477$ 2D MRI slices extracted from the $159$ patients, with $387$ slices being used for training and $90$ slices ($45$ non-deleted and $45$ codeleted) during testing. A test accuracy of $87.7\%$ was obtained.

Studies by the TCGA have established that LGGs can be grouped into three robust molecular classes on the basis of $IDH1/2$ mutations and $1p/19q$ co-deletion. The variants have been reported to differ with respect to tumor margins and internal homogeneity. The $T2$, $FLAIR$ mismatch sign was found to be associated with a survival profile similar to that of the $IDH$-mutant $1p/19q$-non-codeleted glioma subtype \cite{patel17}, and more favorable to that of the $IDH$-wild type gliomas  (which present outcome similar to WHO grade IV glioblastomas).

In this paper we exhaustively investigate the behaviour and performance of ConvNets, with and without transfer learning, for noninvasive studies of gliomas, involving (A) detection and grade prediction (into low- (Grades II and III) and high- (Grade IV) brain tumors (LGG and HGG), and (B) classification of LGG with/without 1p/19q codeletion, from multi-sequence MRI. 

Tumors are typically heterogeneous, depending on cancer subtypes, and contain a mixture of structural and patch-level variability. Prediction of the grade of a tumor may thus be based on either the image patch containing the tumor, or the 2D MRI slice  containing the image of the whole brain including the tumor, or the 3D MRI volume encompassing the full image of the head enclosing the tumor. While in the first case only the tumor patch is necessary as input, the other two cases require the ConvNet to learn to localize the ROI (or VOI) followed by its classification. Therefore, the first case needs only classification while the other two cases additionally require detection or localization. Since the performance and complexity of ConvNets depend on the difficulty level of the problem and the type of input data representation, we introduce three kinds viz. i) Patch-based, ii) Slice-based, and iii) Volume-based data,  from the original MRI dataset by introducing the sliding window concept, and employ these over the two experiments. Three ConvNet models are developed corresponding to each case, and trained from scratch. We also compare two state-of-the-art  ConvNet architectures, viz.  VGGNet \cite{simonyan2014very} and ResNet \cite{he2016deep}, with parameters pre-trained on ImageNet using transfer learning (via fine-tuning).

The main contributions of this research are listed below.
\begin{itemize}
	\item Adaptation of deep learning to Radiomics, for the non-invasive prediction of tumor grade followed determination of the 1p/19q status in Low-Grade Gliomas, from multi-sequence MR images of the brain.
	
	\item Prediction of the grade of brain tumor without manual segmentation of tumor volume, or manual extraction and/or selection of features. 
	
	\item Conceptualization of ``Augmented intelligence'', with the application of deep learning for assisting doctors and radiologists towards decision-making while minimizing human bias and errors.  
	
	\item Development of novel ConvNet architectures viz. PatchNet, SliceNet, and VolumeNet for tumor detection and grade prediction, based on MRI patches, MRI slices, and multi-planar volumetric MR images, respectively.
	  
	\item New framework for applying existing pre-trained deep ConvNets models on multi-channel MRI data using transfer learning. The technique can be further extended to tasks of localization and/or segmentation on different MRI data. 
\end{itemize}

\section*{Results}

The ConvNet models were developed using TensorFlow, with Keras in Python. The experiments were performed on the Intel AI DevCloud platform, having a cluster of Intel Xeon Scalable processors. The quantitative and qualitative evaluation of the results are elaborated below.

\subsection*{Quantitative evaluation}
We use (i) leave-one-patient-out (LOPO), and (ii) holdout (or independent) test dataset for model validation. While only one sample is used for testing in the LOPO scheme, at each iteration, the remaining are employed for training the ConvNets. The process iterated over each patient. Although LOPO test scheme is computationally expensive, it allows availability of more data as required for ConvNets training. LOPO testing is robust and well-suited to our application, with results being generated for each individual patient. Therefore, in cases of misclassification, a patient sample may be further investigated. In holdout or independent testing scheme either a portion of the training data that has never been used for training or a separate test dataset is used during model validation.

Training and validation performance of the three ConvNets were measured using the following two metrics.

$$
Accuracy= \frac{TP + TN}{TP + FP + TN + FN},~~~~~~~~~
F_1 Score = 2 \times \frac{precision \times recall}{precision + recall}.
$$

$Accuracy$ is the most intuitive performance measure and provides the ratio of correctly predicted observations to the total observations. $F_1  Score$ is the weighted average of $Precision$ and $Recall$, which are defined as $\frac{TP}{TP + FP}$ and $\frac{TP}{TP+FN}$, with $TP$, $TN$, $FP$, and $FN$ indicating the numbers of true positive, true negative, false positive and false negative detections. In the presence of imbalanced data one typically prefers $F_1 Score$ over $Accuracy$ because the former considers both false positives and false negatives during computation.

\subsubsection*{Case study-A: Classification of low/high grade gliomas}
The dataset preparation schemes, discussed in Section \textbf{Dataset preparation}, were used to create the three separate training and testing data sets. The ConvNet models PatchNet, SliceNet, VolumeNet, were trained on the corresponding datasets using Stochastic Gradient Descent (SGD) optimization algorithm with learning rate $=0.001$ and momentum = $0.9$, using mini-batches of size $32$ samples generated from the corresponding training dataset. A small part of the training set ($20\%$) was used for validating the ConvNet model after each training epoch, for parameter selection and detection of overfitting. 

Since deep ConvNets entail a large number of free trainable parameters, the effective number of training samples were artificially enhanced using real-time data augmentation -- through some linear transformation such as random rotation ($0^o - 10^o$), horizontal and vertical shifts, horizontal and vertical flips. Also we used Dropout layer with a dropout rate of $0.5$ in the fully connected layers and Batch-Normalization to control the overfitting. After each epoch, the model was validated on the corresponding validation dataset.

\begin{figure*}[]
	\centering
	\includegraphics[width=1.0\textwidth]{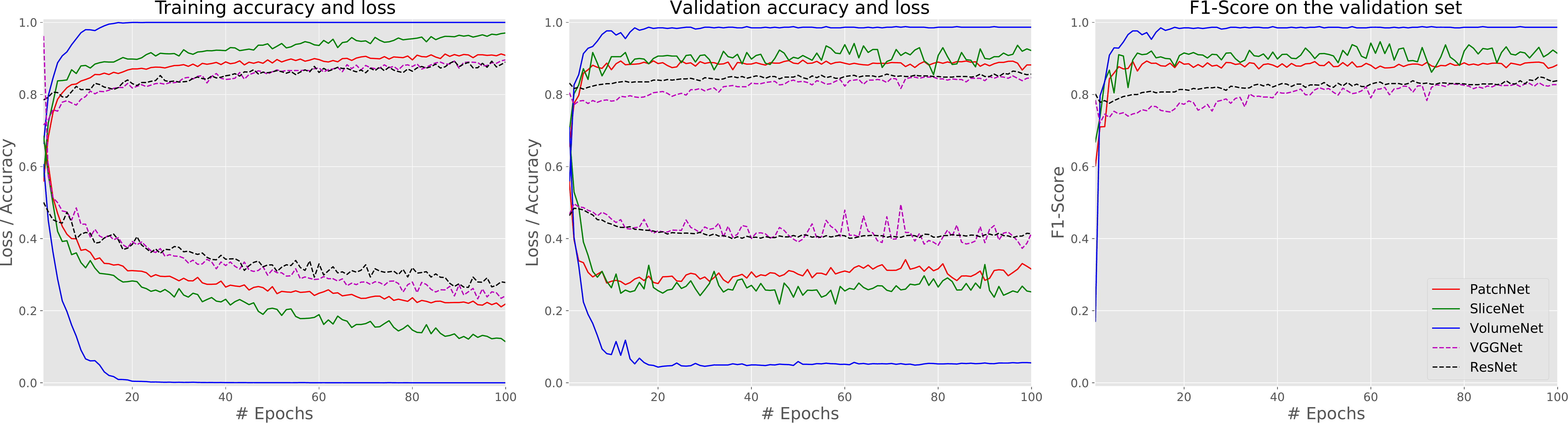}
	\caption{Comparative performance of the networks.}
	\label{fig:fig1}
\end{figure*}

Training and validation $Accuracy$ and loss, and $F_1 Score$ on the validation dataset, are presented in Fig.~\ref{fig:fig1} for the three proposed ConvNets (PatchNet, SliceNet, and VolumeNet), trained from scratch, along with that for the two pre-trained ConvNets (VGGNet, and ResNet) fine-tuned on the TCGA-GBM and TCGA-LGG datasets. The plots demonstrate that VolumeNet gives the highest classification performance during training, reaching maximum accuracy on the training set ($100 \%$) and the validation set ($98\%$) within just $20$ epochs. Although the performance of PatchNet and SliceNet is quite similar on the validation set (PatchNet - $90 \%$, SliceNet - $92 \%$), it is observed that SliceNet achieves better accuracy ($94\%$) on the training set. The performance of two the pre-trained models (VGGNet and ResNet) exhibit similar results, with both achieving around $85\%$ accuracy on the validation set. All the networks reached a plateau after the 50th epoch. This establishes the superiority of the 3D volumetric level processing of VolumeNet.

\textbf{LOPO testing results}: After training, the networks were evaluated on the holdout test set employing majority voting. Each patch or slice from the test dataset was from a single test patient in the LOPO framework, and was categorized as HGG or LGG. The class with maximum number of slices or patches correctly classified was indicative of the grade of the tumor. In case of equal votes the patient was marked as ``ambiguous''. 
\begin{table}[]
	\centering
	\caption{Comparative LOPO test performance}
	\label{tab:tab1}
	{\scriptsize\begin{tabular}{|l|c|c|c|c|}
		\hline
		\multicolumn{1}{|c|}{ConvNets} & Classified   & Misclassified & Ambiguous  & Accuracy          \\ \hline
		PatchNet                       & 242          & 39          & 4          & 84.91 \%          \\ \hline
		SliceNet                       & 257          & 26          & 2          & 90.18 \%          \\ \hline
		\textbf{VolumeNet}             & \textbf{277} & \textbf{8}  & \textbf{0} & \textbf{97.19 \%} \\ \hline
		VGGNet                         & 239          & 40          & 6         & 83.86 \%          \\ \hline
		ResNet                         & 242          & 42          & 1          & 84.91 \%          \\ \hline
	\end{tabular}}
\end{table}
	
The LOPO testing scores are displayed in Table. \ref{tab:tab1}. VolumeNet is observed to achieve the best LOPO test accuracy ($97.19 \%$), with zero ``ambiguous'' cases as compared to the other four networks. SliceNet is also found to provide good LOPO test accuracy ($90.18\%$). Both the pre-trained models show similar LOPO test accuracy as PatchNet. This is interesting because it demonstrates that with a little fine-tuning one can achieve a test accuracy similar to that by the patch-level ConvNet trained from scratch on a specific dataset. Therefore fine-tuning of a few more intermediate layers can lead to very high test scores with little training. 

Table \ref{tab:tab3} compares the proposed ConvNets with existing shallow learning models in literature, used for the same application but requiring additional feature extraction and/or selection from manually segmented ROI/VOI, in terms of classification accuracy. Ref. \cite{banerjee2017fuzzieee} reports the performance achieved by seven standard classifiers, viz. i) Adaptive Neuro-Fuzzy Classifier (ANFC), ii) Naive Bayes (NB), iii) Logistic Regression (LR), iv) Multilayer Perceptron (MLP), v) Support Vector Machine (SVM), vi) Classification and Regression Tree (CART), and vii) $k$-nearest neighbors ($k$-NN), on the BraTS 2015 dataset (a subset of TCGA-GBM and TCGA-LGG datasets) consisting of $200$ HGG and $54$ LGG patient cases, each having $56$ three-dimensional quantitative MRI features manually extracted. On the other hand, the ConvNets leverage the learning capability of deep networks for automatically extracting relevant features from the data.

	\begin{table}[]
		\centering
		\caption{Comparative accuracy of deep and shallow classifiers.}
		\label{tab:tab3}
		{\scriptsize\begin{tabular}{|c|c|l|}
			\hline
			Classifier & Accuracy (\%) & \multicolumn{1}{c|}{Details}                                                                                \\ \hline
			PatchNet   & $84.91$ & \begin{tabular}[c]{@{}l@{}}Trained and tested on MRI patches of size $32 \times 32$, having $3$ ($3 \times 3$) convolution ($8$, $16$, $32$ filters), and \\single FC ($16$ neurons) layers, with $50$ epochs.\end{tabular}                                                                                                       \\ \hline
			SliceNet   & $90.18$ & \begin{tabular}[c]{@{}l@{}}Trained and tested on MRI slices of size $200 \times 200$, having $4$ ($3 \times 3$) convolution ($16$, $32$, $64$, $128$ filters),\\and single FC ($64$ neurons) layers, with $50$ epochs.\end{tabular}                                                                                                         \\ \hline
			\textbf{VolumeNet}  & \textbf{97.19}         & \begin{tabular}[c]{@{}l@{}}Trained and tested on multi-planar MRI slices of size $200 \times 200$ having three parallel ConvNets each with\\$3$ ($3 \times 3$) convolution ($8$, $16$, $32$ filters), and single FC ($32$ neurons) layers, with $10$ epochs.\end{tabular}                                                                                          \\ \hline
			VGGNet     & 83.86         & \begin{tabular}[c]{@{}l@{}}Trained on ImageNet dataset, fine-tuned and tested on MRI slices of size $200 \times 200$
			\end{tabular}                                                                   \\ \hline
			ResNet     & 84.91         & \begin{tabular}[c]{@{}l@{}}Trained on ImageNet dataset, fine tuned and tested on MRI slices of size $200 \times 200$.\end{tabular}                                                                   \\ \hline
			ANFC-LH    & 85.83         & \begin{tabular}[c]{@{}l@{}}Trained on manually extracted $23$ quantitative MRI features, based on 10 fuzzy rules.\end{tabular}                                                                                        \\ \hline
			NB         & 69.48         & \begin{tabular}[c]{@{}l@{}}Trained on manually extracted $23$ quantitative MRI features.\end{tabular}                                                                                        \\ \hline
			LR         & 72.07         & \begin{tabular}[c]{@{}l@{}}Trained on manually extracted $23$ quantitative MRI features based on multinomial logistic regression model with \\a ridge estimator.\end{tabular}           \\ \hline
			MLP        & 78.57         & \begin{tabular}[c]{@{}l@{}}Trained on manually extracted $23$ quantitative MRI features using single hidden layer with $23$ neurons, learning\\rate = $0.1$, momentum = $0.8$.\end{tabular} \\ \hline 
			SVM        & 64.94         & \begin{tabular}[c]{@{}l@{}}Trained on manually extracted $23$ quantitative MRI features, LibSVM with RBF kernel, cost = 1, gamma = 0.\end{tabular}                                         \\ \hline
			CART       & 70.78         & \begin{tabular}[c]{@{}l@{}}Trained on manually extracted $23$ quantitative MRI features using minimal cost-complexity pruning.\end{tabular}                                              \\ \hline
			$k$-NN       & 73.81         & \begin{tabular}[c]{@{}l@{}}Trained on manually extracted $23$ quantitative MRI features, accuracy averaged over scores for $k = 3, 5, 7$.\end{tabular}                                       \\ \hline
		\end{tabular}}
\end{table}

\begin{figure*}[]
	\centering
	\includegraphics[width=1.0\textwidth]{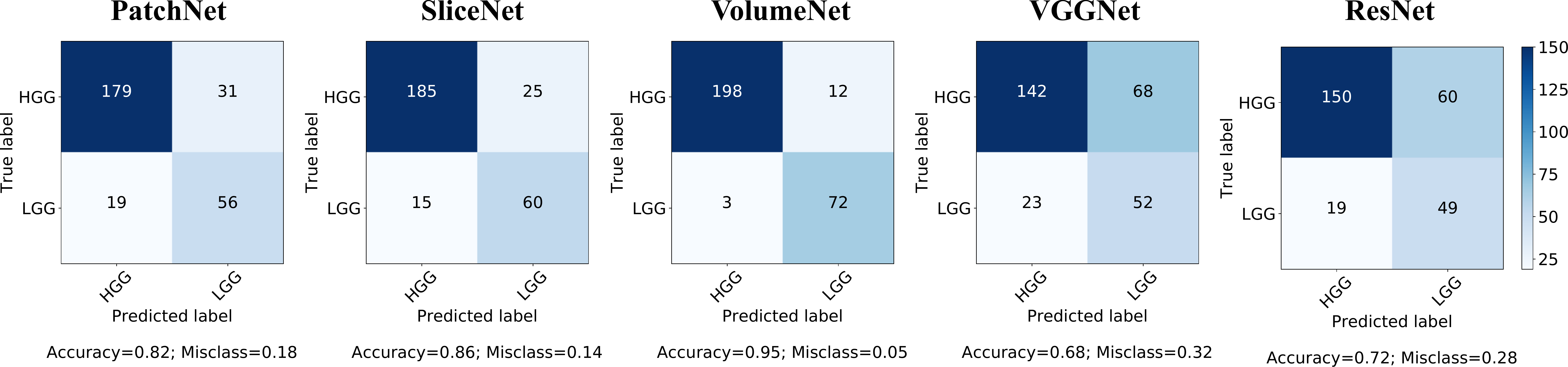}
	\caption{Confusion matrix for classification performance of the five models on MICCAI BraTs 2017 dataset.}
	\label{fig:fig2}
\end{figure*}

\textbf{Testing results on holdout dataset}: Trained networks were also tested on an independent test dataset (MICCAI BraTS 2017 database) as discussed in Section \textbf{Brain tumor data}. The confusion matrix for each of the networks are shown in Fig.~\ref{fig:fig2}. VolumeNet performs the best on the holdout set (accuracy = $95.00\%$). The other two models PatchNet and SliceNet, which were trained from scratch, also demonstrate good classification performance. On the other hand, the fine-tuned models VGGNet and ResNet perform poorly on the independent test dataset. Comparison is made with two recently reported methods\cite{yang2018glioma, cho2018classification}, used for the same problem on same or different datasets. The comparison results are given in Table \ref{tab:tab4}. It is observed that VolumeNet performed the best on the holdout dataset, achieving $7\%$ improvement in the accuracy as compared to state-of-the-art method \cite{cho2018classification} using the same dataset. Note that cross-validation was used in the compared models \cite{yang2018glioma, cho2018classification} for evaluating performance, with training done on a part of the same data on whose holdout portion testing was evaluated. On the other hand, we trained our models on a different dataset (from TCIA) and tested it on the BraTs dataset. This validates the robustness of our models as compared to existing methods.

\begin{table}[]
	\centering
	\caption{Comparative test performance with state-of-the-art methods on holdout dataset.}
	\label{tab:tab4}
	\scriptsize\begin{tabular}{|c|c|c|l|c|}
		\hline
		\multicolumn{2}{|c|}{Model}                                                                                  & Accuracy & Dataset                          & Type                                 \\ \hline
		\multirow{3}{*}{\begin{tabular}[c]{@{}c@{}}Proposed models\\ trained from  scratch\end{tabular}} & PatchNet  & $82\%$   & \multirow{5}{*}{BraTs 2017}      & \multirow{7}{*}{Deep learning based} \\ \cline{2-3}
		& SliceNet  & $86\%$   &                                  &                                      \\ \cline{2-3}
		& VolumeNet & $95\%$   &                                  &                                      \\ \cline{1-3}
		\multicolumn{1}{|l|}{\multirow{2}{*}{Fine-tuned models}}                                         & VGGNet    & $68\%$   &                                  &                                      \\ \cline{2-3}
		\multicolumn{1}{|l|}{}                                                                           & ResNet    & $72\%$   &                                  &                                      \\ \cline{1-4}
		\multirow{2}{*}{Yang et al. \cite{yang2018glioma}}                                            & AlexNet   & $85\%$   & \multirow{2}{*}{Private dataset} &                                      \\ \cline{2-3}
		& GoogLeNet & $90\%$   &                                  &                                      \\ \hline
		\multicolumn{2}{|c|}{Cho et al. \cite{cho2018classification}}                                                    & $88\%$   & Brats 2017                       & Radiomics based                      \\ \hline
	\end{tabular}
\end{table}

\subsubsection*{Case study-B: Classification of LGG with/without 1p/19q codeletion}
\begin{figure}
	\centering
	\begin{minipage}{0.5\textwidth}
		\centering
		\includegraphics[width=1.0\textwidth]{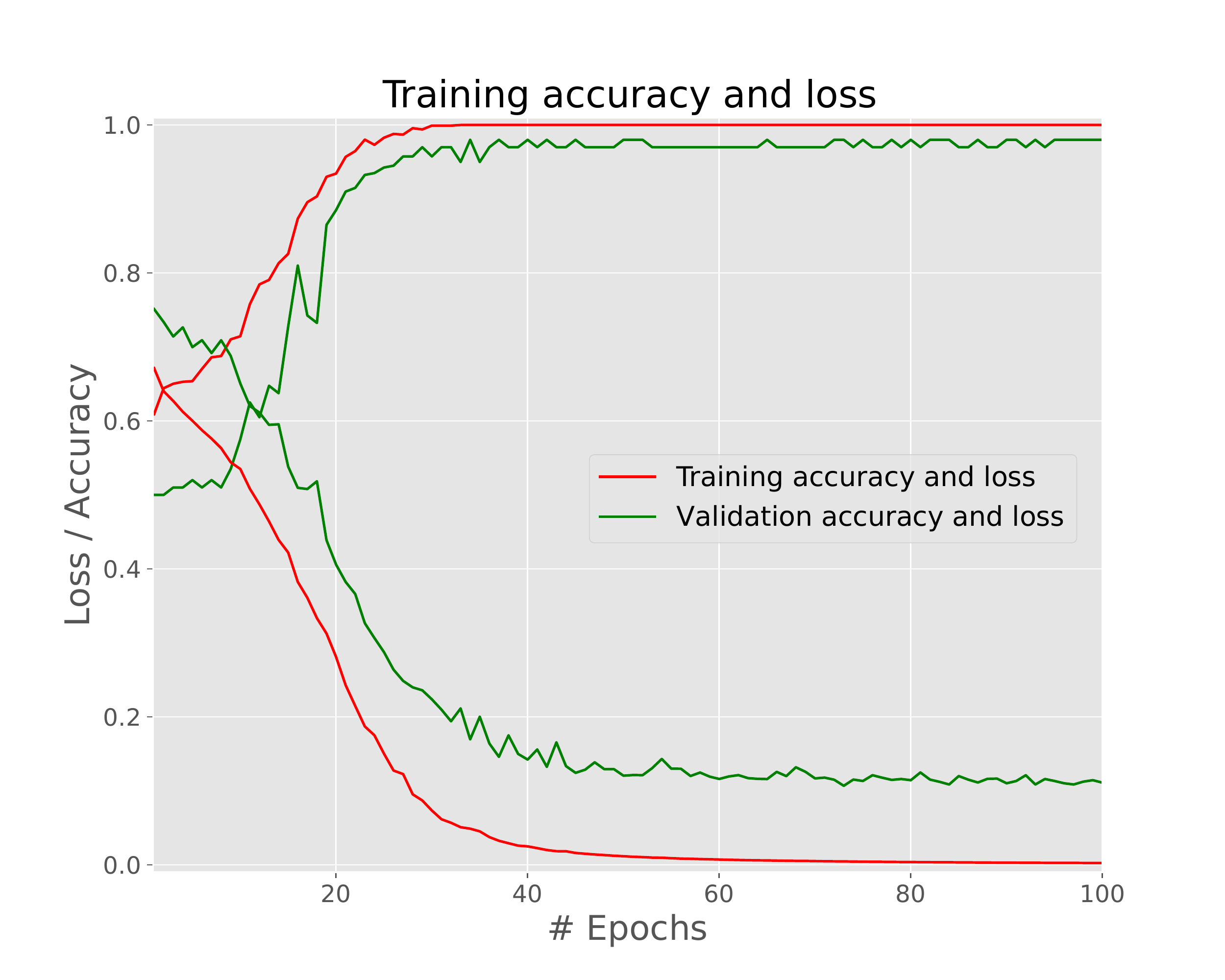}
		\captionof{figure}{Traning performance of VolumeNet for the classification of LGG with/without 1p/19q codeletion.}\label{fig:fig3}
	\end{minipage}
	\begin{minipage}{0.45\textwidth}
		\centering
		\captionsetup{type=table}
		\label{tab:1p19q}
		\begin{tabular}{|c|c|c|}
			\hline
			& VolumeNet & Akkus et al. \cite{akkus2017predicting}\\ \hline\hline
			Sensitivity & 94\%      & 93\%         \\ \hline
			Specificity & 100\%     & 82\%         \\ \hline
			Accuracy    & \textbf{97\%}      & 88\%         \\ \hline
		\end{tabular}
		\captionof{table}{Comparative test performance of VolumeNet on the holdout set.}
	\end{minipage}
\end{figure}

We trained the best performing model, i.e. VolumeNet, for the classification of LGG with/without 1p19q codeletion. The Mayo Clinic database used for the task contains $T1C$ and $T2$ MRI sequences for each patient. VolumeNet was trained on the multi-planar volumetric database, preprocesed from the raw 3D brain volume images as described in Section \textbf{Dataset preparation}. Training performance of the network, in terms of training and validation $Accuracy$ \& $loss$, are presented in Fig.~\ref{fig:fig3}. Comparison was made with a state-of-the-art method\cite{akkus2017predicting}, which is also based on deep learning and uses same dataset. The test performance on the holdout dataset is reported in Table {\color{blue}4}. Here again our model VolumeNet achieves $9\%$ more accuracy than the compared method, over the same dataset. The improvement is due to the incorporation of volumetric information through multi-planar MRI slices.    

\subsection*{Runtime analysis}
The total time required for training each network for $100$ epochs is presented in Table~\ref{tab:tab2}, averaged over several runs. This further corroborates that multi-planar and slice level processing can learn to generalize better than a patch level network, although at the expense of higher computational time.
{
\begin{table}[]
	\centering
	\caption{Comparative training time}
	\label{tab:tab2}
	{\scriptsize\begin{tabular}{|l|l|l|}
		\hline
		ConvNet   & \multicolumn{1}{c|}{\begin{tabular}[c]{@{}c@{}}Time\\ $(Mean \pm SD)$\end{tabular}} & Training type \\ \hline
		PatchNet  & $10.75 \pm 0.05$ min                                                                 & from scratch  \\ \hline
		SliceNet  & $65.95 \pm 0.02$ min                                                                & from scratch  \\ \hline
		VolumeNet	& $132.48 \pm 0.05$ min                                                                                    & from scratch  \\ \hline
		VGGNet	& $8.56 \pm 0.03$    min                                                                                     & fine-tuning   \\ \hline
		ResNet	& $12.14 \pm 0.03$    min                                                                                     & fine-tuning   \\ \hline
	\end{tabular}}
\end{table}}

\subsection*{Qualitative evaluation}
\begin{figure*}[]
	\centering
	\includegraphics[width=1.0\textwidth]{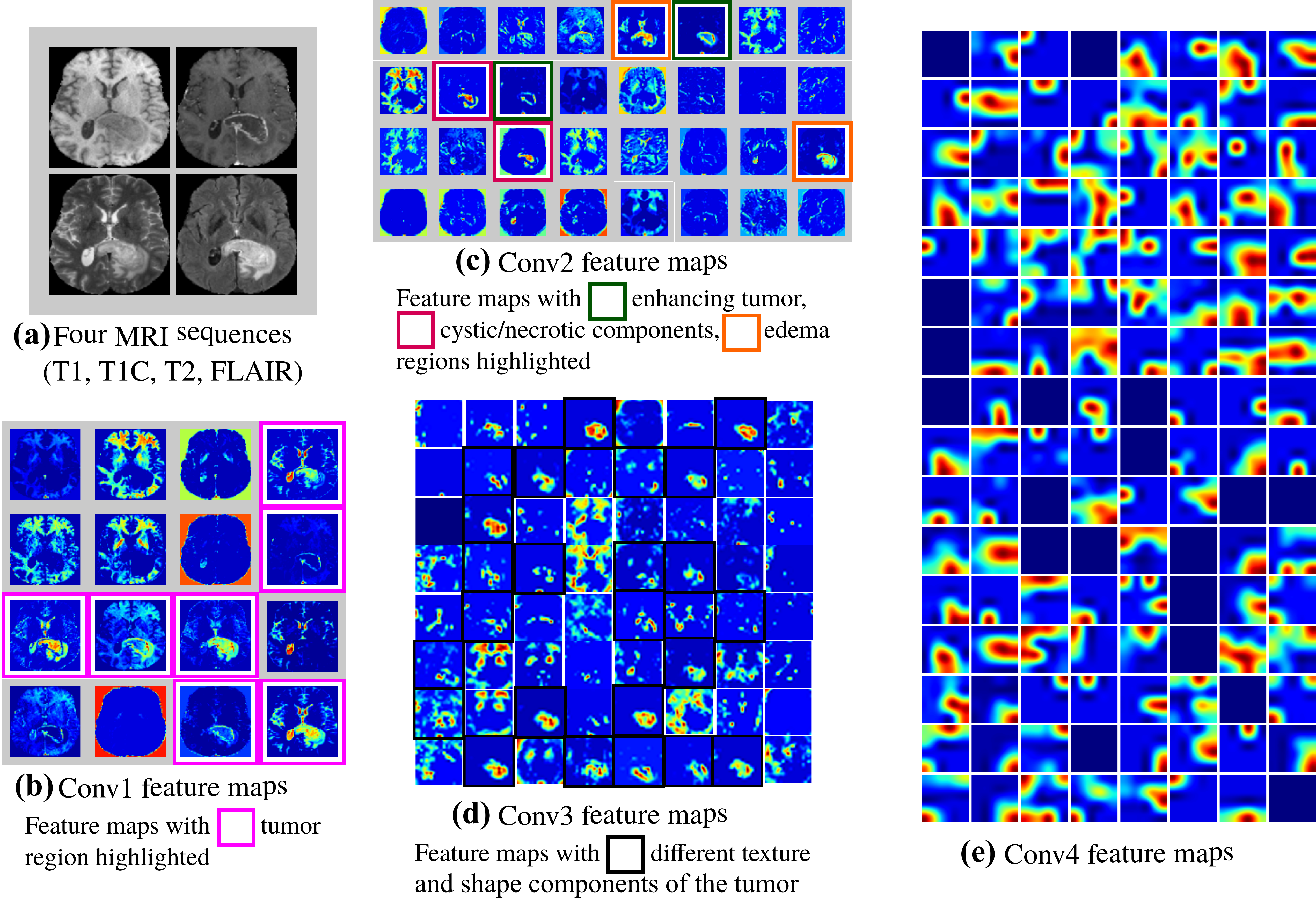}
	\caption{(a) Four sequences of an MRI slice from a sample HGG patient from TCIA~\cite{clark2013cancer} (under a Creative Commons Attribution 3.0 Unported. Full terms at https://creativecommons.org/licenses/by/3.0/). Intermediate layer outputs/feature maps, generated by SliceNet, at diferent levels by (b) Conv1, (c) Conv2, (d) Conv3 and (e) Conv4.}
	\label{fig:fig4}
\end{figure*}

The ConvNets were next investigated through visual analysis of their intermediate layers. The performance of a ConvNet depends on the convolution kernels, which are the feature extractors from the unsupervised learning process. Visualizing the output of any convolution layer can help determine the description of the learned kernels. Fig.~\ref{fig:fig4} illustrates the intermediate convolution layer outputs (after ReLU activation) of the proposed SliceNet architecture on sample MRI slices from an HGG patient.

The visualization of the first convolution layer activations (or feature maps) [Fig.~\ref{fig:fig4}(b)] indicates that the ConvNet has learned a variety of filters to detect edges and distinguish between different brain tissues like white matter (WM), gray matter (GM), cerebrospinal fluid (CSF), skull and background. Most importantly, some of the filters could isolate the ROI (or the tumor); on the basis of which the whole MRI slice may be classified. 

Most of the feature maps generated by the second convolution layer [Fig.~\ref{fig:fig4}(c)] mainly highlight the tumor region and its subregions; like enhancing tumor structures, surrounding cystic/necrotic components and the edema region of the tumor. Thus the filters in the second convolution layer learn to extract deeper features from the tumor by focusing on the ROI (or tumor). 

The texture and shape of the tumor get enhanced in the feature maps generated from the third convolution layer [Fig.~\ref{fig:fig4}(d)]. For example, small, distributed, irregular tumor cells get enhanced (one of the most important tumor grading criteria called ``CE-Heterogeneity'' \cite{chekhun2013tumor}). Finally the last layer [Fig.~\ref{fig:fig4}(e)] extracts detailed information about more discriminating features, by combining these to produce a  clear distinction between images of different types of tumors. 


\section*{Discussion}

An exhaustive study was made to demonstrate the effectiveness of Convolutional Neural networks for non-invasive, automated detection and grading of brain tumors from multi-sequence MR images. Three novel ConvNet architectures were developed for distinguishing between HGG and LGG. Three level ConvNet architectures were designed to handle images at patch, slice and multi-planar modes. This was followed by exploring transfer learning for the same task, by fine-tuning two existing ConvNet models. The scheme for incorporating volumetric tumor information, using multi-planar MRI slices, achieved the best test accuracy of $97.19 \%$ in the LOPO mode and $95\%$ on the holdout dataset for the classification of LGG/HGG (Case study A). In Case study B an accuracy of $97\%$ was obtained for the classification of LGG/HGG (case study A). In case study B an accuracy of $97.00\%$ was obtained for the classification of LGG with/without 1p/19q codeletion status on the holdout dataset. 

Visualization of the intermediate layer outputs/feature maps demonstrated the role of kernels/filters in the convolution layers in automatically learning to detect tumor features closely resembling different tumor grading criteria. It was also observed that existing ConvNets, trained on natural images, performed adequately just by fine-tuning their final convolution layer on the MRI dataset. This investigation allows us to conclude that deep ConvNets could be a feasible alternative to surgical biopsy for brain tumors. 

Diagnosis from histopathological images is also considered to be the ``gold standard" in this domain. HGGs are characterized by the presence of pseudopallisading necrosis (necrotizing cell-devoid region radially surrounded by lined-up tumor cells) and microvascular proliferation (enlarged new blood vessels in the tissue) \cite{mousa15}. The LGGs, on the other hand, exhibit a visual smoothness with cells spread evenly throughout the tissue. Automated glioma grading, through image analysis of the slides, serves to complement the efforts of clinicians in categorizing into low- (Grade II) and high- (Grade III, IV) gliomas (LGG and HGG).

\section*{Methods}

In this section we provide a brief description of the data preparation at three levels of resolution, followed by an introduction to convolutional neural networks and transfer learning.

\subsection*{Brain tumor data}
For the classification of low/high grade gliomas (Case study A), the models were trained on the TCGA-GBM \cite{scarpace9radiology} and TCGA-LGG \cite{pedano2016radiology} datasets downloaded from The Cancer Imaging Archive (TCIA) \cite{clark2013cancer}. The testing was on an independent set, i.e. brain tumor dataset from MICCAI BraTS 2017 \cite{bakas2017advancing, menze2015multimodal} competition containing images of low grade glioma (LGG) and high grade glioma (HGG). The TCGA GBM and LGG datasets consists of $262$ and $199$ samples respectively, whereas the BraTs 2017 database contains $210$ HGG and $75$ LGG samples. Each patient scan has four sequences, encompassing the native ($T1$), post-contrast enhanced $T1$-weighted ($T1C$), $T2$-weighted ($T2$), and $T2$ Fluid-Attenuated Inversion Recovery ($FLAIR$) volumes.

The ConvNet model used for classification of 1p/19q codeletion status in LGG (Case study B) was trained on the brain tumor patient database from Mayo Clinic, containing a total of $159$ LGG patients ($57$ non-deleted and $102$ codeleted) having preoperative postcontrast-$T1$ and $T2$ images downloaded from TCIA\cite{erickson2017data}. A total of $30$ samples ($15$ non-deleted and $10$ codeleted) were randomly selected from the data at the beginning, as a test set, and was never shown to the ConvNet during its training. The remaining $129$ samples were used for training the model.

\begin{figure}[]
	\centering
	\includegraphics[width=0.8\textwidth]{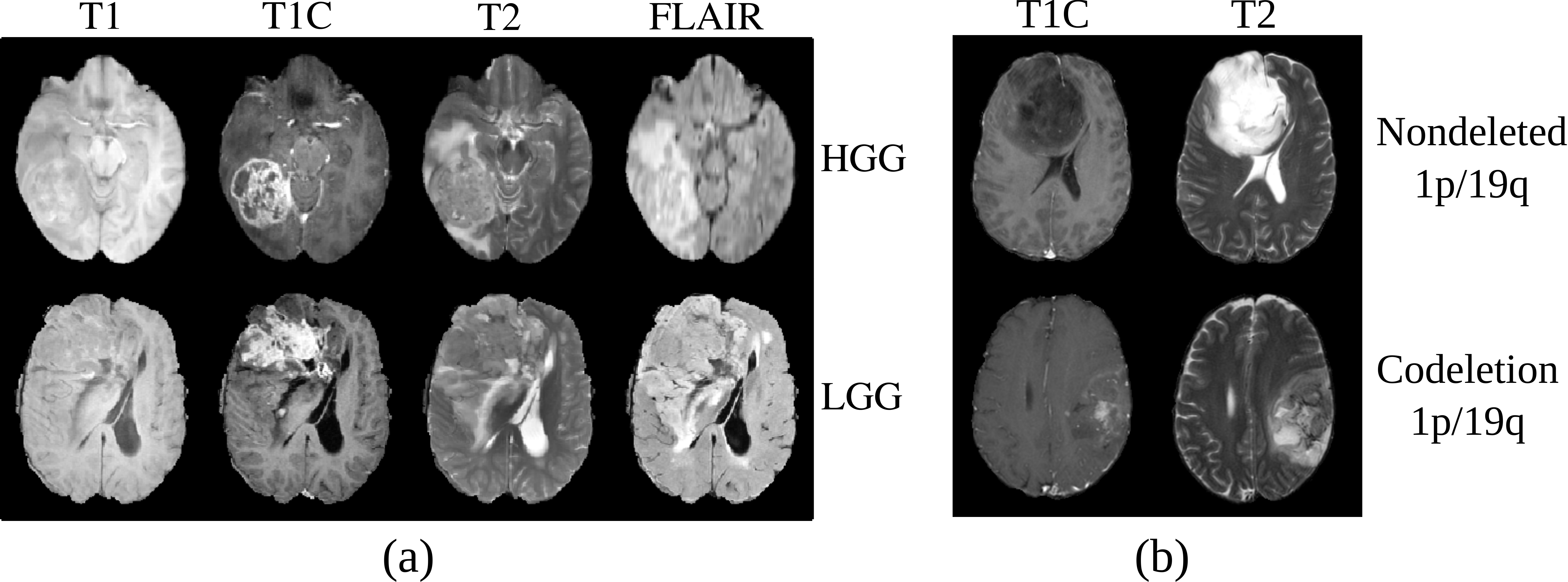}
	\caption{Sample MR image sequences from TCIA \cite{clark2013cancer} (under a Creative Commons Attribution 3.0 Unported. Full terms at https://creativecommons.org/licenses/by/3.0/) of (a) low/high grade gliomas, and (b) low-grade glioma with and without 1p/19q codeletion.}
	\label{fig:fig5}
\end{figure}

Sample images of the two glioma grades (LGG/HGG), and LGG with and without 1p/19q codeletion are shown in Fig.~\ref{fig:fig5}(a), (b), respectively. It can be observed from the figure that it is very hard to discriminate in each case, based only on the phenotypes visible to the human eye. Hence abstract features learned by the deep layers of a ConvNet are expected to be helpful in noninvasively differentiating between them. Besides, the use of large public domain datasets can allow more clinical impact as compared to controlled and dedicated prospective image acquisitions.

The datasets were aligned to the same anatomical template, skull-stripped, bias field corrected and interpolated to $1mm^3$ voxel resolution.

\subsection*{Dataset preparation}\label{dataset_preparation}
Although the TCGA-GBM and TCGA-LGG datasets consist MRI volumes, we cannot propose a 3D ConvNet model for the classification problem; mainly because the dataset has only $262$  HGG and $199$ LGG patients data, which is considered as inadequate to train a 3D ConvNet with a huge number of trainable parameters. Another problem with the dataset is its imbalanced class distribution i.e. about $35.72\%$ of the data comes from the LGG class. Therefore we formulate 2D ConvNet models based on the MRI patches (encompassing the tumor region) and slices, followed by a multi-planar slice-based ConvNet model that incorporates the volumetric information as well. 

Applying ConvNet directly on the MRI slice could require extensive downsampling, thereby resulting in loss of discriminative details. The tumor can be lying anywhere in the image and can be of any size (scale) or shape. Therefore classifying the tumor grade from patches is easier, because the ConvNet learns to localize only within the extent of the tumor in the image.  Thereby the ConvNet needs to learn only the relevant details without getting distracted by irrelevant details. However it may lack spatial and neighborhood details of the tumor, which may adversely influence grade prediction. Although classification based on the 2D slices and patches often achieves good accuracy, the incorporation of volumetric information from the dataset can enable the ConvNet to perform better.

Along these lines, we propose schemes to prepare three different sets viz.  (i) patch-based, (ii) slice-based, and (iii) multi-planar volumetric, from the TCIA datasets. 

\subsubsection*{Patch-based dataset} The slice with the largest tumor region is first identified. Keeping this slice in the middle, a set of slices before and after it are considered for extracting 2D patches containing the tumor regions using a bounding-box.  This bounding-box is marked, corresponding to each slice, based on the ground truth image. The enclosed image region is then extracted. 

\begin{figure}[tbp]
	\centering
	\includegraphics[width=0.4\textwidth]{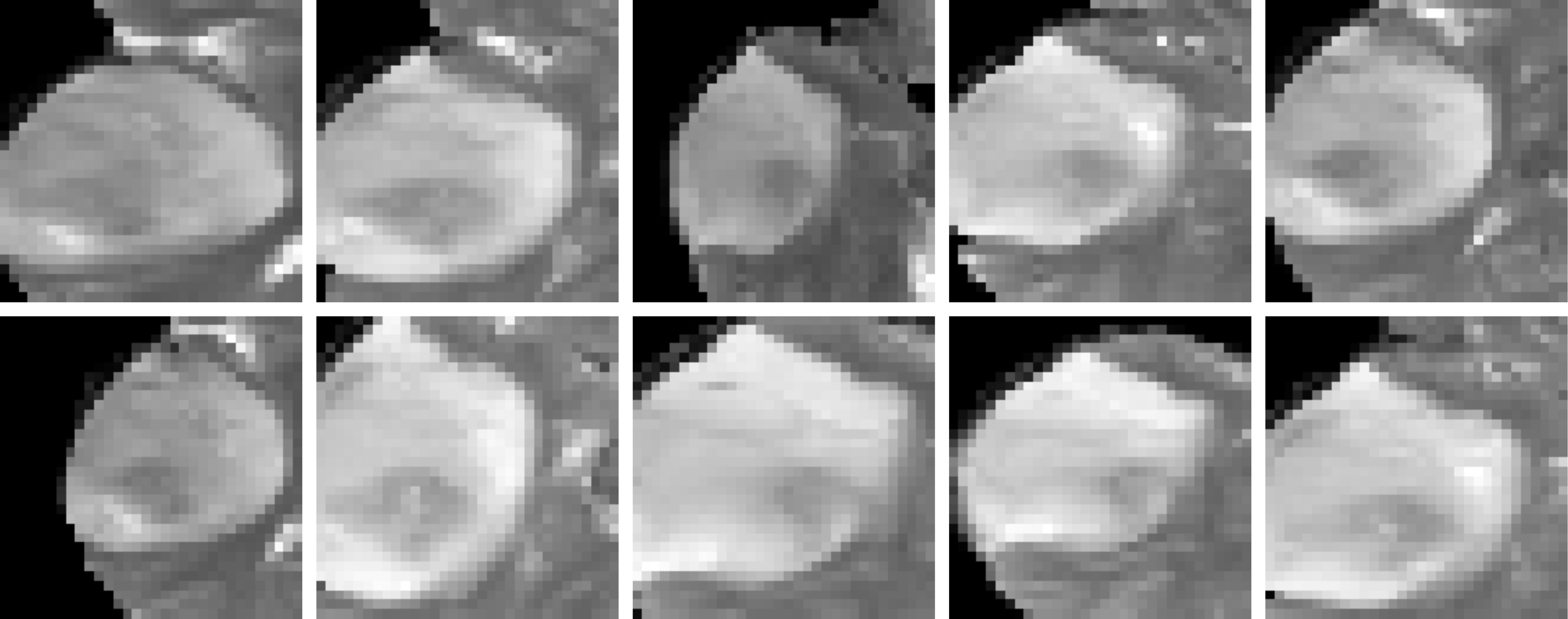}
	\caption{Ten T2-MR patches extracted from contiguous slices from an LGG patient from TCIA \cite{clark2013cancer} (under a Creative Commons Attribution 3.0 Unported. Full terms at https://creativecommons.org/licenses/by/3.0/).}
	\label{fig:fig6}
\end{figure}

We use a set of $20$ slices for extracting the patches. In case of MRI volumes from HGG (LGG) patients, four (ten) 2D patches [with a skip over $5$ ($2$) slices] are extracted for each of the MR sequences. Therefore a total of $210 \times 4 = 840$ HGG and $75 \times 10 = 750$ LGG patches, with four channels each, constitute this dataset. Although the classes are still not perfectly balanced, this ratio is found to be good enough in the scenario of this enhanced dataset. 

In spite of significant dissimilarity visible between contiguous MRI slices at a global level, there may be little difference exhibited at the patch level. Therefore patches extracted from contiguous MRI slices look similar, particularly for LGG cases. Fig.~\ref{fig:fig6} depicts a set of $10$ patches extracted from contiguous MR slices of an LGG patient. This can lead to overfitting in the  ConvNet. To overcome this problem we introduce a concept of static augmentation by randomly changing the perfect bounding-box coordinates by a small amount ($\in\{-5, 5\}$ pixels) before extracting the patch. This results in improved learning and convergence of the network. 
 
\subsubsection*{Slice-based dataset}
Complete 2D slices, with visible tumor region, are extracted from the MRI volume. The slice with the largest tumor region, along with a set of $20$ slices before and after it, are extracted from the MRI volume in a sequence similar to that of the patch-based approach. While for HGG patients $4$  (with a skip over $5$) slices are extracted, in the case of LGG patients $10$ (with a skip of $2$) slices are used.

\subsubsection*{Multi-planar volumetric dataset} Here 2D MRI slices are extracted along all three anatomical planes, viz. axial ($X$-$Z$ axes), coronal ($Y$-$X$ axes), and sagittal ($Y$-$Z$ axes), in a manner similar to that described above. 

\subsection*{Convolutional neural networks}\label{ConvNets}
Convolutional Neural Networks (ConvNets) can automatically learn low-level, mid-level and high-level abstractions from input training data in the form of convolution filter weights, that get updated during the training process by backpropagation. The inputs percolating through the network are the responses of convoluting the images with various filters. These filters act as detectors of simple patterns like lines, edges, corners, from spatially contiguous regions in an image. When arranged in many layers, the filters can automatically detect prevalent patterns while blocking irrelevant regions. Parameter sharing and sparsity of connection are the two main concepts that make ConvNets easier to train with a small number of weights as compared to dense fully connected layers. This reduces the chance of overfitting, and enables learning translation invariant features. Some of the important concepts, in the context of ConvNets are next discussed.

\subsubsection*{Layers} The fundamental constituents of a ConvNet consist of the input, convolution, activation, pooling and fully-connected layers. The input layer receives a multi-channel brain MRI patch/slice denoted by $I \in \mathbb{R}^{m \times w \times h}$, where $m$ is the number of channels, $w$ and $h$ represent the resolution of the image. The convolutional layer takes the image or feature maps as input, and performs the convolution operation between the input and each of the filters to generate a set of activation maps. The output feature map dimension, from a convolution layer, is  calculated as $w_{out}/h_{out} = \frac{(w_{in}/h_{in} - F + 2P)}{S} + 1$,	where $w_{in}$ and $h_{in}$ are the width and height of the input image, $w_{out}$ and $h_{out}$ are the width and height of the effective output. Here $P$ denotes the input padding, the stride $S=1$, and $F$ is the kernel size of the neurons in a particular layer. Output responses of the convolution and fully connected layers pass through some nonlinear activation function, such as a Rectified Linear Unit (ReLU) \cite{glorot2011deep}, for transforming the data. ReLU, defined as $f(a) = max(0, a)$, is a popular activation function for deep neural networks due to its computational efficiency and reduced likelihood of vanishing gradient. The pooling layer follows each convolution layer to typically reduce computational complexity by downsampling of the convoluted response maps. Max pooling enables selection of the maximum feature response in local neighborhoods, and thereby enhances translation invariance. The features learned through a series of convolutional and pooling layers are eventually fed to a fully-connected layer, typically a Multilayer Perceptron. Additional layers like Batch-Normalization \cite{ioffe2015batch} reduce initial covariate shift. Dropout \cite{srivastava2014dropout} is used as regularizer to learn a better representation of the data.   

\subsubsection*{Loss} The cost function for the ConvNets is chosen as binary cross-entropy (for a two-class problem) as
\begin{equation}
L_C = -\frac{1}{n}\sum_{i=1}^{n}\left\{y_i \log(f_i) + (1 - y_i) \log(1 - f_i)\right\},
\end{equation}
where $n$ is the number of samples, $y_i$ is the true label of a sample and $f_i$ is its  predicted label.

\subsection*{Transfer learning}\label{transfer_learning}
Typically the early layers of a ConvNet learn low-level image features, which are applicable to most vision tasks. The later layers, on the other hand, learn high-level features which are more application-specific. Therefore, shallow fine-tuning of the last few layers is usually sufficient for transfer learning. A common practice is to replace the last fully-connected layer of the pre-trained ConvNet with a new fully-connected layer, having as many neurons as the number of classes in the new target application. The rest of the weights, in the remaining layers, of the pre-trained network are retained. This corresponds to training a linear classifier with the features generated in the preceding layer. However, when the distance between the source and target applications is significant then one may need to induce deeper fine-tuning. This is equivalent to training a shallow neural network with one or more hidden layers. An effective strategy \cite{tajbakhsh2016} is to initiate fine-tuning from the last layer, and then incrementally include deeper layers in the tuning process until the desired performance is achieved.

\section*{ConvNets for Brain tumor Grading \label{brain_tumor_grading_using_ConvNets}}
This section introduces the three ConvNet architectures, trained on the three level brain tumor MR data sets, along with a brief description of the fine tuning of existing models.

\subsection*{Three level architectures}
\begin{figure*}[tbp]
	\centering
	\includegraphics[width=0.85\textwidth]{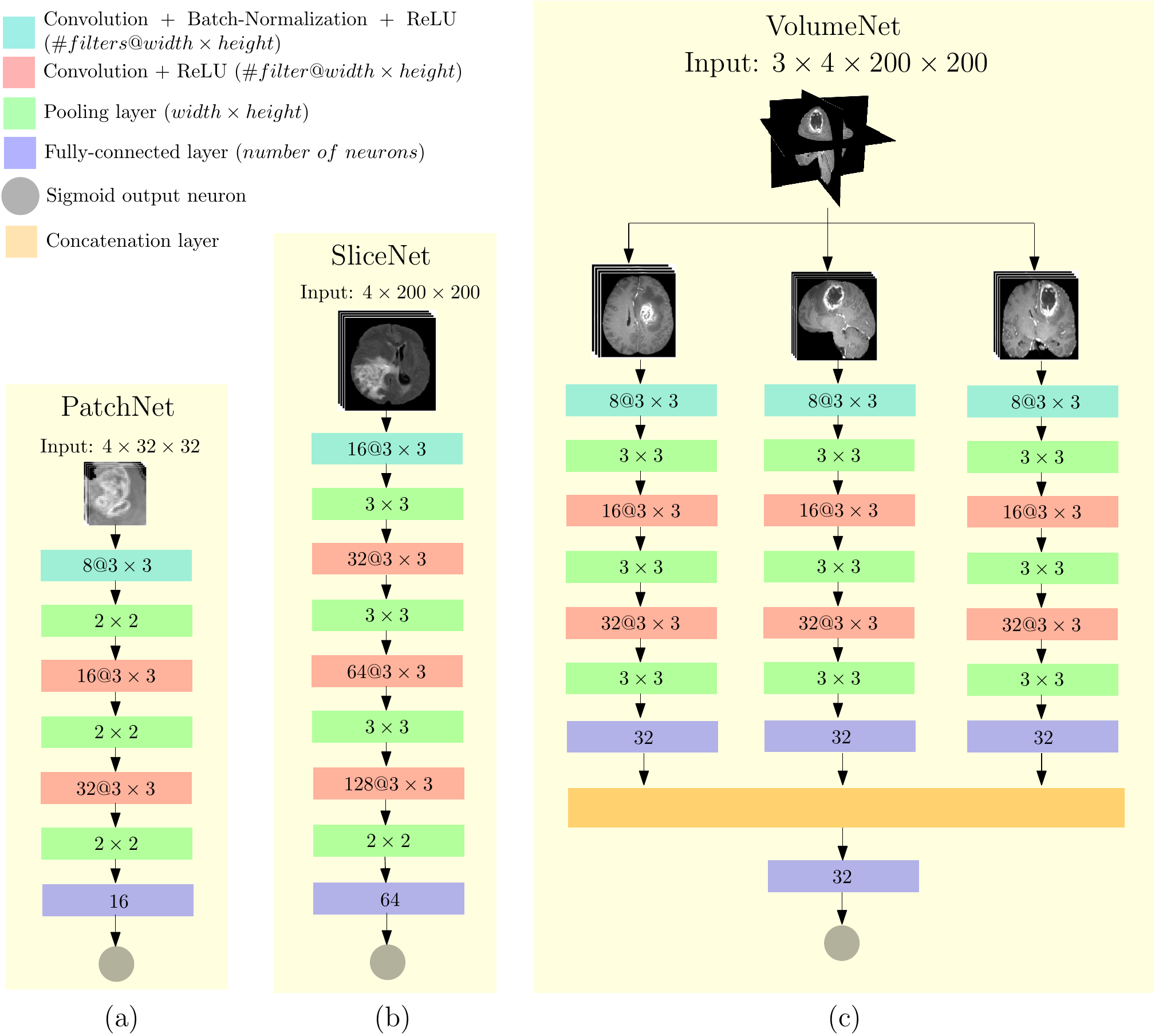}
	\caption{Three level ConvNet architectures (a) PatchNet, (b) SliceNet, and (c) VolumeNet, sample MRIs are from TCIA database \cite{clark2013cancer} (under a Creative Commons Attribution 3.0 Unported. Full terms at https://creativecommons.org/licenses/by/3.0/).}
	\label{fig:fig7}
\end{figure*}

We propose three ConvNet architectures, named PatchNet, SliceNet, and VolumeNet, which are trained from scratch on the three datasets prepared as detailed in Section \textbf{Dataset preparation}. This is followed by transfer learning and fine-tuning of these networks. The ConvNet architectures are illustrated in Fig.~\ref{fig:fig7}. PatchNet is trained on the patch-based dataset, and provides the probability of a patch belong to HGG or LGG. SliceNet gets trained on the slice-based dataset, and predicts the probability of a slice being from HGG or LGG. Finally VolumeNet is trained on the multi-planar volumetric dataset, and predicts the grade of a tumor from its 3D representation using the multi-planar 3D MRI data. 

As reported in literature \cite{szegedy2015going}, smaller size convolutional filters produce better regularization due to the smaller number of trainable weights; thereby allowing construction of deeper networks without losing too much information in the layers. We use filters of size $(3 \times 3)$ for our ConvNet architectures. A greater number of filters, involving deeper convolution layers, allows for more feature maps to be generated. This compensates for the decrease in size of each feature map caused by ``valid'' convolution and pooling layers. Due to the complexity of the problem and bigger size of the input image, the SliceNet and VolumeNet architectures are deeper as compared to the PatchNet.

\subsection*{Fine-tuning}

Pre-trained VGGNet (16 layers), and ResNet (50 layers) architectures, trained on the ImageNet dataset, are employed for transfer learning. Even though ResNet is deeper than VGGNet, the model size of ResNet is substantially smaller due to the usage of global average pooling rather than fully-connected layers. Transferring from the non-medical to the medical image domain is achieved through fine-tuning of the last convolutional block of each model, along with the fully-connected layer (top-level classifier). Fine-tuning of a trained network is achieved by retraining on the new dataset, while involving very small weight updates. The adopted procedure is outlined below.

\begin{itemize}
	\item Instantiate the convolutional base of the model and load its pre-trained weights.
	\item  Replace the last fully-connected layer of the pre-trained ConvNet with a new fully-connected layer, having single neuron with sigmoid activation.
	\item Freeze the layers of the model up to the last convolutional block.
	\item Finally retrain the last convolution block and the fully-connected layers using Stochastic Gradient Descent (SGD) optimization algorithm with a very slow learning rate.
\end{itemize} 

Since the base models were trained on RGB images, and accept single input with three channels, we train and test them on the slice-based dataset involving three MR sequences ($T1C$, $T2$, $FLAIR$). The $T1C$ sequence was found to perform better than $T1$, when used in conjunction with $T2$ and $FLAIR$. Although running either of these two models from scratch is very expensive, particularly on CPU, the concept of fine-tuning just the last few layers could be easily accomplished.

\bibliography{mybib}

\section*{Acknowledgements}
This research is supported by the IEEE Computational Intelligence Society Graduate Student Research Grant 2017.

\noindent S. Banerjee acknowledges the support provided to him by the Intel Corporation, through the Intel AI Student Ambassador Program.

\noindent S. Mitra acknowledges the support provided to her by the Indian National Academy of Engineering, through the INAE Chair Professorship.

\noindent This publication is an outcome of the R\&D work undertaken in a project with the Visvesvaraya PhD Scheme of Ministry of Electronics \& Information Technology, Government of India, being implemented by Digital India Corporation.

\section*{Author contributions statement}
S.B. conceived the experiment(s),  S.B. conducted the experiment(s), S.B. analysed the results.  S.B, S.M, F.M, and S.R reviewed the manuscript. 

\section*{Additional Information}
\subsection*{Competing interests}
The author(s) declare no competing interests.
\end{document}